\documentclass{article} 
\usepackage{iclr2023_mlrs_workshop_mine,times}
\pdfoutput=1

\usepackage{amsmath,amsfonts,bm}

\def\eqref#1{equation~\ref{#1}}

\def\1{\bm{1}}

\DeclareMathAlphabet{\mathsfit}{\encodingdefault}{\sfdefault}{m}{sl}
\SetMathAlphabet{\mathsfit}{bold}{\encodingdefault}{\sfdefault}{bx}{n}

\usepackage{hyperref}
\usepackage{url}
\usepackage{amsmath}
\usepackage{cleveref}

\title{Evaluation challenges for geospatial ML}

\author{Esther Rolf \\%
Harvard University
}

\newcommand{\loc}[0]{\ell}
\newcommand{\evalset}[0]{\mathcal{S}_{\textrm{eval}}}
\newcommand{\pop}[0]{\mathcal{D}}

\iclrfinalcopy 
\begin{document}

\maketitle

\begin{abstract}
As geospatial machine learning models and maps derived from their predictions are increasingly used for downstream analyses in science and policy, it is imperative to evaluate their accuracy and applicability. 
Geospatial machine learning has key distinctions from other learning paradigms, and as such, the correct way to measure performance of spatial machine learning outputs has been a topic of debate. 
In this paper, I delineate unique challenges of model evaluation for geospatial machine learning with global or remotely sensed datasets, 
culminating
in concrete takeaways to improve  evaluations of geospatial model performance. 
%
%

\end{abstract}

\section{Motivation}
\label{sec:intro}

Geospatial machine learning (ML), for example with remotely sensed data, is being used across consequential domains, including  
public health \citep{nilsen2021review,draidi2022spatial}
conservation  \citep{sofaer2019development}, food security \citep{nakalembe2018characterizing}, and wealth estimation 
\citep{jean2016combining,chi2022microestimates}. 
By both their use and their very nature, geospatial predictions have a purpose beyond model benchmarking;
mapped data are to be read, scrutinized, and acted upon.
Thus, it is critical to rigorously and comprehensively evaluate how well a predicted map represents the state of the world it is meant to reflect, or how well a spatial ML model performs across the many conditions in 
which it might be used.
%

Unique structures in remotely sensed and geospatial data complicate or even invalidate use of traditional ML evaluation procedures. 
Partially as a result of misunderstandings of these complications, 
the stated performance of several geospatial models and predictive maps has come into question \citep{
fourcade2018paintings, ploton2020spatial}.
This in turn has sparked disagreement on what the ``right'' evaluation procedure  is. 
With respect to a certain set of spatial evaluation methods (described in \S\ref{sec:spatial_crossval}), one is  jointly presented with the arguments that ``spatial cross-validation is essential in preventing overoptimistic model performance''  \citep{meyer2019importance} and ``spatial cross-validation methods have
no theoretical underpinning and should not be used for assessing map accuracy'' \citep{wadoux2021spatial}. 
That both statements can simultaneously hold reflects the importance of using a diverse set of evaluation methods tailored to the many ways in which a geospatial ML model might be used.
%

In this paper, I situate the challenges of geopsatial model evaluation in the perspective of an ML researcher, synthesizing prior work across ecology, geology, statistics, and machine learning. 
I aim in part to disentangle key factors that complicate effective evaluation of model and map performance.
First and foremost, evaluation procedures should be designed to measure as closely as possible the quantity or phenomena they are intended to assess (\S\ref{sec:map_accuracy_vs_model_performance}).
After the relevant performance measures are established, considerations can be made about 
what is feasible with the available data (\S\ref{sec:data_section}).
%
With all of this in mind, possible evaluation procedures (\S\ref{sec:spatial_eval_methods}) can be compared and tailored to the task at hand.
Recognizing the interaction of these  distinct but related steps  exposes opportunities to improve geospatial performance assessment, both in individual studies and more broadly (\S\ref{sec:discussion_oppurtuntities}).

\section{Map accuracy and model performance: Contrasting views}
\label{sec:map_accuracy_vs_model_performance} 
Estimating accuracy 
indices and  corresponding  
uncertainties 
of geospatial predictions is essential to reporting geospatial ML performance (\S\ref{subsec:map_accuracy}), especially when prediction maps will be used for downstream analyses 
or policy decisions.
At the same time, the potential value of a geospatial ML model likely extends beyond that of a single mapped output (\S\ref{sec:prediction_regimes}).
Delineating the (possibly many) facets of desired model and map use is key to measuring geospatial ML performance
 (\S\ref{sec:contrasting_views}).


\subsection{Map accuracy as a population parameter to be estimated}
\label{subsec:map_accuracy}
%
%

%
Establishing notation we will use throughout, let $\hat{y}(\loc)$ denote a model's predicted value  at location $\loc$, and $y(\loc)$ the reference, or ``ground truth'' value (which we assume can be measured). 
To calculate a \textbf{map accuracy index as a population parameter} 
for accuracy index $F$ is to calculate
$
A(\pop) = F(\{ (\hat{y}(\loc), y(\loc))\}_{\loc \in \pop})
$
where $\pop$ is the \textbf{target population} of map use (e.g. all (lat, lon) pairs in a global grid, or all administrative units in a set of countries). 
Examples of common $F$ 
include root mean squared error,
and
area under the ROC curve, among many others \citep{maxwell2021accuracy}. 

Typically, one only has a limited set of values $y$ for locations in an evaluation set $\loc \in \evalset$ from which to compute a statistic
$ \hat{A}(\evalset) $
to estimate $A(\pop)$.
%
%
%
%
\citet{wadoux2021spatial} discuss the value of using a design-independent probability sample for design-based estimation of $A$ (in contrast, model-based estimation makes statistical  assumptions about the data \citep{brus2021statistical}).
Here a \textbf{design-independent sample} is one collected independently of the model training process. 
A \textbf{probability sample} is one for which every location in $\pop$ has 
a positive probability of appearing in $\evalset$, and these probabilities are known for all $\ell \in \evalset$ 
%
(see, e.g. \citet{lohr2021sampling}).
\citet{wadoux2021spatial} emphasize that 
when $\evalset$ is a design independent probability sample from population $\pop$,
design-based inference can be used to estimate $A(\mathcal{\pop})$ with $\hat{A}(\evalset)$, 
\emph{regardless of the prediction model or distribution of training data}.

Computing statistically valid estimates of map accuracy indices is clearly a key component of reporting overall geospatial ML model performance.
It is often important 
to understand how accuracy and uncertainty in predictions vary across sub-populations $\pop_{r_1},  \pop_{r_2} \ldots \subset \pop$ (such as administrative regions or climate zones \citep{meyer2022machine}).
If \textbf{local accuracy} indexes $A(\pop_{r_1}),  A(\pop_{r_2}) \ldots$ are low in certain sub-regions, this could expose concerns about fairness or model applicability. 
%

\subsection{Model performance extends beyond map accuracy}
\label{sec:prediction_regimes}
Increasingly, geospatial ML models are designed with the goal of being used \emph{outside} of the regions where training labels are available. Models trained with globally available remotely sensed data might be used to ``fill in'' spatial gaps common to other data modalities (\S\ref{sec:data_availability}). The goals of \textbf{spatial generalization}, \textbf{spatial extrapolation} or \textbf{spatial domain adaption} can take different forms: e.g. applying a model trained with data  from one region 
to a wholly new region, 
or using data from a few clusters or subregions to extend predictions across the entire region. 
When spatial generalizability is desired, performance should be assessed specifically with respect to this goal (\S\ref{sec:spatial_eval_methods}).

%

%

While spatial generalization is a key component of performance for many geospatial models,
it too is just one facet of geospatial model performance.
Proposed uses of geospatial ML models and their outputs include estimation of natural or causal parameters \citep{NBERw30861}, and reducing autocorrelation of prediction residuals in-sample \citep{song2022three}. 
Other important facets of geospatial ML performance  are model interpretability \citep{brenning2022spatial} and usability, including the resources required to train, deploy and maintain models \citep{rolf2021generalizable}.  
%

%

\subsection{Contrasting perspectives on performance assessment}
\label{sec:contrasting_views}
The differences between estimating map accuracy as a population parameter (\S\ref{subsec:map_accuracy})
and assessing a model's performance in the conditions it is most likely to be used 
(\S\ref{sec:prediction_regimes})
are central to one of the
discrepancies introduced in \S\ref{sec:intro}.
%
%
\citet{meyer2019importance, ploton2020spatial, meyer2022machine} state concerns in light of numerous ecological studies applying non-spatial validation techniques with the explicit purpose of spatial generalization. 
They rightly caution that when data exhibit spatial correlation (\S\ref{sec:spatial_data_structures}), non-spatial validation methods will almost certainly over-estimate predictive performance in these use cases.
\citet{wadoux2021spatial}, in turn,
argue that performance metrics from spatial validation methods will
not necessarily tell you anything about $A$ as a population parameter.

A second discrepancy between these two perspectives hinges on what data is assumed to be available (or collectable).
%
While there are some major instances of probability samples being collected for evaluation of global-scale maps  \citep{boschetti2016stratified,stehman2021validation}, this is far from standard standard  in geospatial ML studies \citep{maxwell2021accuracy}. 
More often, datasets are created ``by merging all data available from different sources'' \citep{meyer2022machine}.
Whatever the intended use of a geospatial model, the availability of and structures within geopsatial and remotely sensed data must be contended with in order to reliably evaluate any sort of performance.

\section{Structures and patterns in spatial and remotely sensed data}
\label{sec:data_section}
Geospatial and remotely sensed data exhibit distinct structures. 
For example, the chosen extent and scale of a spatial prediction unit
 ($\ell$ in \S\ref{sec:map_accuracy_vs_model_performance}) has important implications for the design, use, and evaluation of geospatial ML models, evidenced by the phenomena of  ``modifiable areal unit problem'' and ``ecological fallacy'' \citep{haining2009special,nikparvar2021machine,yuan2022embedding}.
Here, I focus on two key factors of geospatial data that affect 
the validity of
geospatial ML evaluation methods.

\subsection{Spatial structures: (Auto)correlation and covariate shift}
\label{sec:spatial_data_structures}
One key phenomena exhibited by many geopsatial data is that values of a variable (e.g. tree canopy height) are often correlated across locations. Formally, for random process $Z$, the  \textbf{spatial autocorrelation function} is defined as 
$
R_{ZZ}(\loc_i, \loc_j) = \mathbb{E}[Z(\loc_i) Z(\loc_j)]  / \sigma_i \sigma_j 
$, where $\sigma_i, \sigma_j$ are the standard deviations associated with $Z(\loc_i),Z(\loc_j)$.
%
%
%
For geospatial variables, we might expect $R_{ZZ}(\loc_i, \loc_j) > 0$ when $\loc_i$ and $\loc_j$ are closer together, namely that values of $Z$ at nearby points tend to be closer in value.
%
The degree of spatial autocorrelation in data 
can be assessed with statistics such as Moran's $I$ and  Geary's $C$, and semi-variogram analsyes (see, e.g. \citep{gaetan2010spatial}). 

Spatial autocorrelations and  correlations between predictor and label variables can be an important source of structure to leverage in geospatial ML models \citep{rolf2020post,klemmer2021auxiliary}, yet they also present challenges.
%
Models can ``over-rely'' on spatial correlations in the data, leading to over-estimated  accuracy despite poor spatial generalization performance. 
%
Overfitting to spatial relationships in training data is of particular concern in the when data distributions differ between training regions and regions of use. 
Such \textbf{covariate shifts} are common in geospatial data, e.g. across climate zones, or spectral shifts in satellite imagery  \citep{tuia2016domain,hoffimann2021geostatistical}.

Presence of spatial correlations or domain shift alone do not invalidate  assessing map accuracy with a probability sample  (\S\ref{subsec:map_accuracy}). 
%
%
However, when evaluation is limited to existing data, issues of data availability and representivity can amplify the challenges of geospatial model evaluation.

\subsection{Availability, quality, and representivity of geospatial evaluation data}
\label{sec:data_availability}

Many geospatial datasets exhibit gaps in coverage or quality of data.
\citet{meyer2022machine} evidence trends of geographic clustering around research sites primarily in a small number of countries, across
three datasets used for global mapping in ecology. 
%
%
\citet{oliver2021global} find geographical bias in coverage of species distribution data  aggregated from 
field observation, sensors measurements, and citizen science efforts. 
\citet{burke2021using}
note that the frequency at which nationally representative data on agriculture, population, and economic factors are collected varies widely across the globe.  
%
%
While earth observation data such as satellite imagery have comparatively much higher coverage across time and space \citep{burke2021using}, coverage of commercial products has been shown to be biased toward regions of high commercial value \citep{dowman2017global}.


Filling in data gaps is goal for which  geospatial ML can be transformative (\S\ref{sec:prediction_regimes}), yet these same gaps complicate model evaluation. 
When training data are clustered in small regions, this can affect our ability to train a high-performing model. 
When \emph{evaluation data} are clustered in small regions, this affects our ability to evaluate geospatial ML model performance at all. 




\section{Spatially-aware evaluation methods: A brief overview}

In \S\ref{sec:data_section}, we established that geospatial data  generally exhibit spatial correlations and 
data gaps, even when target use areas $\pop$ are small.
It is well documented that 
calculating accuracy indices with non-spatial validation methods (e.g. standard k-fold cross-validation) will generally \emph{over-estimate}  performance in such settings. 
%
%
%
%
Spatially-aware evaluation methods can control the spatial distribution of training and validation set points to better simulate conditions of intended model use.

\label{sec:spatial_eval_methods}
\subsection{Spatial cross-validation methods}
\label{sec:spatial_crossval}

Several spatial cross-validation methods have been proposed 
that reduce spatial dependencies between
train set points $\ell \in \mathcal{S}_{\textrm{train}}$ from evaluation set points $\ell \in \mathcal{S}_{\textrm{eval}}$.
%
Spatial cross-validation methods
typically stratify training and evaluation instances by larger geographies \citep{roberts2017cross, valavi2018blockcv}
e.g. existing boundaries, spatial blocks, or automatically generated clusters.
Buffered cross-validation methods (such as spatial leave-one-out \citep{le2014spatial}, leave-pair out \citep{airola2019spatial} and k-fold cross validation \citep{pohjankukka2017estimating}) control the minimum distance from any training point to any evaluation point.
%
In addition to evaluating model performance, 
spatial cross-validation has also been suggested as a way to to improve model selection and parameter estimation in geospatial ML \citep{meyer2019importance,schratz2019hyperparameter, roberts2017cross}.

While separating $\ell \in \mathcal{S}_{\textrm{train}}$ 
from $\ell \in \mathcal{S}_{\textrm{eval}}$ can  reduce the amount of correlation between training and evaluation data, a spatial split also induces a higher degree of spatial extrapolation to the learning setup and potentially reduces variation in the evaluation set labels.
%
As a result, it is possible for spatial validation methods to systematically \emph{under-report} performance, especially in interpolation regimes \citep{roberts2017cross, wadoux2021spatial}. 
%
In a different flavor from the evaluation methods above, \citet{mila2022nearest} propose to match the distribution of nearest neighbor distances between train and evaluation set points to the corresponding distances between train set and target use area. 


\subsection{Other spatially-aware validation evaluation methods}
\label{subsec:other_spatial_eval}
%
%
When the intended use of a geospatial model is to generate predictions outside the training distribution, it is critical to test the model's ability to generalize across different conditions.
For example, studies have varied the amount of spatial extrapolation required by changing parameters of the spatial validation setups in \S\ref{sec:spatial_crossval} , e.g. with buffered leave one out \citep{ploton2020spatial,brenning2022spatial} and checkerboard designs \citep{roberts2017cross,rolf2021generalizable}. 
\citet{jean2016combining} assess extrapolation ability across pairs of countries by iteratitvely training in one region and evaluating performance in another. \citet{rolf2021generalizable} find that the distances at which a geospatial model has extrapolation power can differ substantially depending on the properties of the prediction variable.
%

%

It is always critical to put the reported performance of geospatial ML models in context.
Visualizing the spatial distributions of predictions and error residuals can help expose overreliance on spatially correlated predictors \citep{meyer2019importance} and sub-regions with low local model performance.
Comparing performance to that of a baseline model built entirely on spatial predictors can contextualize the value-add of a new geospatial model \citep{fourcade2018paintings, rolf2021generalizable}.
%
%

\section{Taking stock: Considerations and opportunities}
\label{sec:discussion_oppurtuntities}
Comprehensive reporting of performance is critical for geospatial ML methods, 
especially as stated gains in research progress make their way to maps and decisions of real-world consequence. 
Evaluating performance  of geospatial models is especially challenging in the face of spatial correlations and limited availability or representivity of data. 
This means non-spatial data splits are generally unsuitable for geospatial model evaluation with most existing datasets. 
Spatially-aware validation methods are an important indicator of model performance including spatial generalization; 
however, 
they generally do not provide valid statistical estimates of prediction map accuracy. 
%
%
%
This brings us to end with three key opportunities for improving the landscape of geospatial ML evaluation: 

\textbf{Opportunity 1: Invest in evaluation data} to measure map accuracy and overall performance of geospatial models.
When remote annotations are appropriate, labeling tools (e.g. \citet{robinson2022fast}) can facilitate the creation of probability-sampled evaluation datasets.
Data collection and aggregation efforts can focus on filling existing geospatial data gaps \citep{paliyam2021street2sat} or simulating real-world prediction conditions like covariate or domain shift \citep{koh2021wilds}.

\textbf{Opportunity 2: Invest in evaluation frameworks} to precisely and transparently and report performance and valid uses of a geospatial ML model (\`{a} la 
``model cards'' \citep{mitchell2019model}).
This includes improving spatial baselines, expanding methods for reporting
uncertainty over space, and delineating ``areas of applicability" for geospatial models, e.g. as in \citet{meyer2022machine}.

%

\textbf{Opportunity 3:} If the available data and evaluation frameworks are insufficient, \textbf{explain the limitations of what types of performance can be evaluated}. Distinguish between performance measures that estimate a statistical parameter and those that indicate potential skill for a possible use case. 

\subsubsection*{Acknowledgments}
Esther Rolf was supported by the Harvard Data Science Initiative (HDSI) and the Center for Research
on Computation and Society (CRCS). Thank you to Konstantin Klemmer,  Caleb Robinson, and Jessie Finocchiaro for feedback on earlier drafts of this work.
%
%
\bibliography{spatial_references}
\bibliographystyle{iclr2023_conference}

\end{document}